\documentclass[sigconf,nonacm]{acmart}

\usepackage{amsmath}
\usepackage{booktabs}
\usepackage{microtype}
\usepackage{graphicx}
\usepackage{enumitem}
\usepackage{tikz}
\usepackage{algorithm}
\usepackage{algorithmic}
\usetikzlibrary{arrows.meta, positioning, shapes.geometric, fit, backgrounds}

\acmConference[AIDEEDS'26]{AI-Driven Energy Efficiency in Dynamic Systems}{June 22, 2026}{Banff, Canada}
\acmDOI{}
\acmISBN{}

\renewcommand\footnotetextcopyrightpermission[1]{}

\begin{document}

\title{Adaptive Reservoir Computing for Multi-Scenario Chaotic System Forecasting}

\author{Shadmehr Zaregarizi}
\affiliation{%
  \institution{Politecnico di Torino}
  \city{Turin}
  \country{Italy}
}
\email{shadmehr.zaregarizi@studenti.polito.it}

\author{Khashayar Yavari}
\affiliation{%
  \institution{Politecnico di Torino}
  \city{Turin}
  \state{TO}
  \country{Italy}
}
\email{khashayar.yavari@studenti.polito.it}

\renewcommand{\shortauthors}{Zaregarizi and Yavari}

\begin{abstract}
We present an adaptive reservoir computing framework for the CTF-4-Science Lorenz
benchmark, which evaluates machine learning models across twelve distinct tasks
spanning five qualitatively different scenarios: baseline forecasting, noisy signal
reconstruction, forecasting under noise, few-shot learning, and parametric
generalization. Rather than applying a uniform inference strategy, we tailor the
training and prediction procedure of Echo State Networks (ESNs) to the specific
demands of each evaluation scenario. Our key contributions are fourfold: (1) exact
reservoir state synchronization that eliminates warmup approximation error in
short-time prediction; (2) histogram-guided candidate selection that directly
optimizes the long-time ergodic evaluation metric; (3) multi-seed reservoir search
for few-shot regimes with severely limited training data; and (4) sequential
multi-sequence training that resolves state-distribution mismatch in parametric
generalization tasks. The proposed framework achieves a score of 74.91 on the
public benchmark leaderboard, demonstrating that carefully adapted reservoir
computing constitutes a competitive and computationally efficient approach for
diverse chaotic system modeling challenges.
\end{abstract}


\maketitle

%-----------------------------------------------------------------------
\section{Introduction}
%-----------------------------------------------------------------------

The Lorenz attractor~\cite{lorenz1963deterministic}, introduced in 1963, is among
the most extensively studied chaotic dynamical systems. Its defining feature sensitive
dependence on initial conditions renders long-horizon trajectory prediction
fundamentally bounded by the Lyapunov time (${\approx}1.1$ time units for standard
parameters), beyond which individual forecasts inevitably diverge. Nevertheless,
chaotic attractors possess rich statistical structure: their ergodic invariant
measures remain reproducible and predictable in a distributional sense even when
pointwise trajectory accuracy is lost. Modeling both the short-time fidelity and
the long-time statistical properties of chaotic systems is a core challenge in
scientific machine learning, with direct applications in weather forecasting,
turbulent flow modeling, epidemiological dynamics, and nonlinear control
systems~\cite{pathak2018model,vlachas2020backpropagation}.

The CTF-4-Science framework~\cite{wyder2025ctf} addresses this challenge through
standardized, reproducible benchmarks for Scientific ML. The Lorenz competition
defines twelve evaluation metrics (E1--E12) across nine dataset pairs, probing five
distinct modeling capabilities: accurate baseline forecasting, denoising and
reconstruction, robust prediction under noise, few-shot generalization from minimal
observations, and parametric generalization to unseen system regimes. This breadth
demands algorithms that are not only accurate but fundamentally adaptable across
qualitatively different data conditions.

Reservoir computing~\cite{jaeger2001echo,maass2002real}---and specifically Echo
State Networks (ESNs)---offers a compelling framework for chaotic system modeling.
ESNs employ a fixed, randomly initialized dynamical system (the \emph{reservoir})
as a nonlinear feature extractor, training only a linear readout via ridge
regression. This design yields efficient optimization, avoids vanishing gradients,
and has been shown to accurately replicate the invariant measures of low-dimensional
chaotic attractors~\cite{pathak2017using,lu2018attractor}.

In this paper we develop an \emph{adaptive} ESN framework that matches the
inference strategy to each evaluation scenario. We demonstrate that principled
synchronization and scenario-aware strategy selection allows a single reservoir
architecture to compete effectively across all nine challenge pairs without any
gradient-based optimization or deep learning infrastructure.

%-----------------------------------------------------------------------
\section{Related Work}
%-----------------------------------------------------------------------

\paragraph{Reservoir Computing for Chaos.}
Jaeger and Haas~\cite{jaeger2004harnessing} first demonstrated that ESNs can learn
to predict chaotic time series by exploiting the echo state property. Pathak
et al.~\cite{pathak2017using} showed that ESNs can replicate not only the
short-time trajectory of the Lorenz system but also reproduce its invariant
measure and Lyapunov exponents from data alone. This was extended to
high-dimensional spatiotemporal chaos in~\cite{pathak2018model}, where reservoir
computers predicted Kuramoto-Sivashinsky dynamics up to eight Lyapunov times
using only local information. Lu et al.~\cite{lu2018attractor} demonstrated
attractor reconstruction from partial observations. More recently,
Vlachas et al.~\cite{vlachas2020backpropagation} compared reservoir computing
against LSTM and backpropagation-through-time methods on chaotic benchmarks,
finding that ESNs remain competitive despite their simpler training procedure.

\paragraph{Few-Shot and Multi-Task Time Series Learning.}
Learning dynamical systems from limited observations is an active research area.
Standard approaches include physics-informed neural networks~\cite{raissi2019physics},
transfer learning across parameter regimes, and Bayesian methods for uncertainty
quantification. The CTF-4-Science benchmark~\cite{wyder2025ctf} formalizes this
multi-scenario challenge, enabling systematic comparison of approaches across
diverse data regimes including both data-rich and data-scarce settings.

\paragraph{Scientific ML Benchmarks.}
Standardized benchmarks are essential for measuring progress in scientific ML.
The Common Task Framework for Science~\cite{wyder2025ctf} provides evaluation
protocols that separate short-time accuracy, long-time statistical fidelity, and
reconstruction quality---metrics that are often conflated in standard time series
benchmarks. The same framework has subsequently been extended to other scientific
domains, including seismic wavefield modeling~\cite{yermakov2025seismic} and
nuclear fission/fusion surrogate models~\cite{riva2026ctf4nuclear}, demonstrating
the generality of the CTF methodology. Our work is the first to systematically
address all five scenario types of the Lorenz benchmark using a single reservoir
computing architecture.

%-----------------------------------------------------------------------
\section{Problem Formulation}
%-----------------------------------------------------------------------

\subsection{Lorenz System}
The three-dimensional Lorenz system evolves according to:
\begin{align}
\dot{x} &= \sigma(y - x), \label{eq:lorenz1}\\
\dot{y} &= x(\rho - z) - y, \label{eq:lorenz2}\\
\dot{z} &= xy - \beta z, \label{eq:lorenz3}
\end{align}
with fixed $\sigma = 10$, $\beta = 8/3$, and $\rho$ varying across evaluation pairs
to test generalization. All trajectories are discretized at $\Delta t = 0.05$ using
fourth-order Runge-Kutta integration.

\subsection{Evaluation Tasks}
Table~\ref{tab:pairs} summarizes the nine evaluation pairs and their scenarios.
Three evaluation metrics are applied:
\begin{itemize}[leftmargin=*,noitemsep,topsep=2pt]
  \item \textbf{Short-time} (E1, E7, E9, E11, E12): RMSE over the first $k{=}20$
        predicted time steps.
  \item \textbf{Long-time} (E2, E4, E6, E8, E10): Histogram L2 error comparing
        marginal distributions over 41 bins evaluated along 500 dominant modes.
  \item \textbf{Reconstruction} (E3, E5): L2 error between denoised predictions
        and the noise-free reference trajectory.
\end{itemize}
All scores are normalized to $[-100, 100]$, where 100 denotes perfect performance.

\begin{table*}[t]
\caption{The nine evaluation pairs, their scenarios, training data, and our scenario-specific inference strategy.}
\label{tab:pairs}
\small
\setlength{\tabcolsep}{6pt}
\begin{tabular}{cllllp{4.5cm}}
\toprule
\textbf{Pair} & \textbf{Evals} & \textbf{Scenario} & \textbf{Train} & \textbf{Data Type} & \textbf{Strategy} \\
\midrule
1 & E1, E2  & Baseline forecasting        & 10{,}000 steps & Clean              & ESN + exact terminal state $\mathbf{r}^*$ \\
2 & E3      & Medium-noise reconstruction  & 10{,}000 steps & Noisy              & Teacher-forced ESN (streaming) \\
3 & E4      & Medium-noise forecasting     & 10{,}000 steps & Noisy              & Histogram-guided candidate selection \\
4 & E5      & High-noise reconstruction    & 10{,}000 steps & Noisy              & Teacher-forced ESN (streaming) \\
5 & E6      & High-noise forecasting       & 10{,}000 steps & Noisy              & Histogram-guided candidate selection \\
6 & E7, E8  & Few-shot clean learning      & 100 steps      & Clean              & Multi-seed ESN sweep (30 seeds) \\
7 & E9, E10 & Few-shot noisy learning      & 100 steps      & Noisy              & Multi-seed ESN sweep (30 seeds) \\
8 & E11     & Parametric interpolation     & 3$\times$10{,}000 steps & Clean           & Sequential multi-sequence training \\
9 & E12     & Parametric extrapolation     & 3$\times$10{,}000 steps & Clean           & Sequential multi-sequence training \\
\bottomrule
\end{tabular}
\end{table*}

%-----------------------------------------------------------------------
\section{Methodology}
%-----------------------------------------------------------------------

\subsection{Echo State Network Architecture}

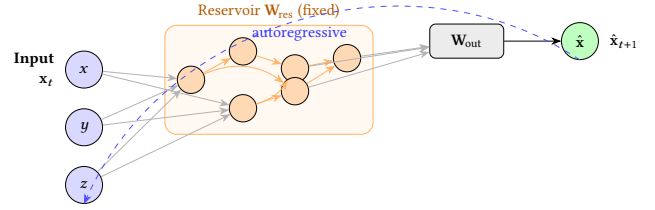
\begin{figure}[t]
\centering
\resizebox{\columnwidth}{!}{%
\begin{tikzpicture}[
  node distance=0.6cm and 1.0cm,
  >=Stealth,
  every node/.style={font=\small},
  inputnode/.style={circle, draw, fill=blue!15, minimum size=0.65cm, inner sep=1pt},
  resnode/.style={circle, draw, fill=orange!30, minimum size=0.48cm, inner sep=0pt},
  outnode/.style={circle, draw, fill=green!25, minimum size=0.65cm, inner sep=1pt},
  readout/.style={rectangle, draw, fill=gray!15, rounded corners=3pt,
                  minimum width=1.3cm, minimum height=0.6cm, inner sep=3pt},
]
\node[inputnode] (x1) {$x$};
\node[inputnode, below=0.35cm of x1] (x2) {$y$};
\node[inputnode, below=0.35cm of x2] (x3) {$z$};

\node[resnode, right=1.3cm of x1, yshift=-0.18cm] (r1) {};
\node[resnode, right=0.42cm of r1, yshift=0.5cm]  (r2) {};
\node[resnode, right=0.42cm of r1, yshift=-0.5cm] (r3) {};
\node[resnode, right=0.42cm of r2, yshift=-0.3cm] (r4) {};
\node[resnode, right=0.42cm of r3, yshift=0.3cm]  (r5) {};
\node[resnode, right=0.42cm of r4, yshift=0.18cm] (r6) {};

\begin{scope}[on background layer]
  \node[draw=orange!60, fill=orange!5, rounded corners=6pt,
        fit=(r1)(r2)(r3)(r4)(r5)(r6), inner sep=6pt,
        label={[font=\small,orange!70!black]above:Reservoir $\mathbf{W}_{\!\text{res}}$ (fixed)}] (resbox) {};
\end{scope}

\node[readout, right=1.2cm of r6, yshift=0.3cm] (wout) {$\mathbf{W}_{\!\text{out}}$};
\node[outnode, right=1.0cm of wout] (out) {$\hat{\mathbf{x}}$};

\foreach \i in {x1,x2,x3}
  \foreach \r in {r1,r3}
    \draw[->, gray!60, thin] (\i) -- (\r);

\draw[->, orange!60, thin] (r1) -- (r2);
\draw[->, orange!60, thin] (r2) -- (r4);
\draw[->, orange!60, thin] (r3) -- (r5);
\draw[->, orange!60, thin] (r4) -- (r6);
\draw[->, orange!60, thin] (r5) -- (r6);
\draw[->, orange!60, thin] (r1) to[bend left=30] (r5);
\draw[->, orange!60, thin] (r3) to[bend right=20] (r4);

\foreach \r in {r4,r5,r6}
  \draw[->, gray!60, thin] (\r) -- (wout);

\draw[->] (wout) -- (out);

\draw[->, dashed, blue!70, bend right=50]
  (out.south) to node[below, font=\small, blue!80]{autoregressive} (x3.south);

\node[font=\small, left=0.08cm of x1, align=right] {\textbf{Input}\\$\mathbf{x}_t$};
\node[font=\small, right=0.08cm of out] {$\hat{\mathbf{x}}_{t+1}$};
\end{tikzpicture}
}
\caption{ESN architecture. The reservoir $\mathbf{W}_{\!\text{res}}$ is fixed after
random initialization; only $\mathbf{W}_{\!\text{out}}$ is trained. During forecasting,
outputs are fed back as inputs (dashed arrow).}
\label{fig:esn}
\end{figure}

We employ leaky-integrator ESNs with reservoir state update:
\begin{equation}
\mathbf{r}_{t+1} = (1-\alpha)\,\mathbf{r}_t
  + \alpha\,\tanh\!\bigl(\mathbf{W}_{\!\text{res}}\,\mathbf{r}_t
  + \mathbf{W}_{\!\text{in}}\,\mathbf{x}_t\bigr),
\label{eq:esn}
\end{equation}
where $\mathbf{r}_t \in \mathbb{R}^N$ is the reservoir state, $\alpha$ the leaking
rate, $\mathbf{W}_{\!\text{res}} \in \mathbb{R}^{N \times N}$ a sparse random
reservoir matrix scaled to spectral radius $\rho_W$, and
$\mathbf{W}_{\!\text{in}} \in \mathbb{R}^{N\times 3}$ a dense random input matrix.
The linear readout $\hat{\mathbf{x}}_{t+1} = \mathbf{r}_t\,\mathbf{W}_{\!\text{out}}$
is obtained via ridge regression:
\begin{equation}
\mathbf{W}_{\!\text{out}} =
  \bigl(\mathbf{R}^\top\mathbf{R} + \lambda\mathbf{I}\bigr)^{-1}
  \mathbf{R}^\top\mathbf{Y},
\label{eq:ridge}
\end{equation}
where $\mathbf{R}$ collects post-washout reservoir states and $\mathbf{Y}$ the
corresponding one-step-ahead targets. Figure~\ref{fig:esn} illustrates the
architecture. Standard hyperparameters: $N{=}1000$, $\rho_W{=}1.2$,
sparsity $= 0.1$, $\alpha{=}0.3$, input scaling $= 0.5$, $\lambda{=}10^{-7}$,
washout $= 500$.

\subsection{Exact Reservoir State Synchronization (Pair 1)}
\label{sec:sync}

After teacher-forced training over $T$ time steps, the terminal reservoir state
$\mathbf{r}^* = \mathbf{r}_{T-1}$ is exactly synchronized to the final training
trajectory point $\mathbf{x}_{T-1}$. Standard practice discards this state and
re-warms the reservoir from a short training tail, introducing avoidable
approximation error. We store $\mathbf{r}^*$ during \texttt{fit()} and initialize
all forecasts directly from this exact state. Since short-time accuracy (E1) is
evaluated only over the first 20 steps, maximal IC synchronization directly
improves performance at the critical evaluation horizon.

\subsection{Teacher-Forced Reconstruction (Pairs 2, 4)}

For reconstruction, noisy observations $\tilde{\mathbf{x}}_t$ drive the reservoir
via teacher forcing at every step; the readout produces the denoised estimate
$\hat{\mathbf{x}}_{t+1} = \mathbf{r}_t\,\mathbf{W}_{\!\text{out}}$. This projects
each noisy input onto the learned attractor manifold, exploiting the ESN's implicit
knowledge of the clean dynamics. The approach is fully streaming ($O(N)$ memory),
enabling efficient processing of 10,000-step trajectories without storing the full
state matrix.

\subsection{Histogram-Guided Candidate Selection (Pairs 3, 5)}

For long-time forecasting from noisy initial conditions, trajectory-level accuracy
is unachievable due to noise-induced IC uncertainty. We instead optimize directly
for the long-time histogram metric by generating $C{=}20$ candidate forecasts from
randomly perturbed warmup sequences ($\sigma_\epsilon = 3{\times}10^{-6}$) and
selecting the candidate minimizing the empirical histogram L2 error:
\begin{equation}
\hat{c} = \arg\min_{c \in \{1,\dots,C\}}
  \sum_{i=1}^{3}\sum_{b=1}^{B}
  \bigl(h^{\text{train}}_{i,b} - h^{(c)}_{i,b}\bigr)^2,
\label{eq:hist_select}
\end{equation}
where $h_{i,b}$ is the normalized histogram count for coordinate $i$, bin $b$,
with $B{=}41$. This aligns the optimization target directly with the evaluation criterion.

\subsection{Multi-Seed Few-Shot Learning (Pairs 6, 7)}

With only 100 training steps (88 effective samples after washout $= 10$), a standard
large ESN overfits or fails to explore the attractor adequately. We use smaller
reservoirs ($N{=}300$, $\lambda{=}10^{-4}$) and sweep 30 random reservoir seeds,
selecting the best forecast by histogram score. The seed sweep exploits diversity of
random reservoir connectivities, effectively forming an ensemble over random feature
maps. For Pair~7 (noisy few-shot), teacher-forced training on noisy observations
provides implicit regularization.

\subsection{Sequential Multi-Sequence Training (Pairs 8, 9)}

The parametric generalization pairs provide three 10,000-step training trajectories
plus 100-step initialization sequences at test parameter values. A critical subtlety
arises: if each training sequence is processed independently from
$\mathbf{r}_0{=}\mathbf{0}$, the fitted $\mathbf{W}_{\!\text{out}}$ reflects
transient states from independent initializations. At inference, however, the
reservoir is chained through the initialization sequence from the terminal training
state---a different state distribution causing mismatch and potential instability
(observed as reservoir states exceeding attractor bounds, $z > 100$).

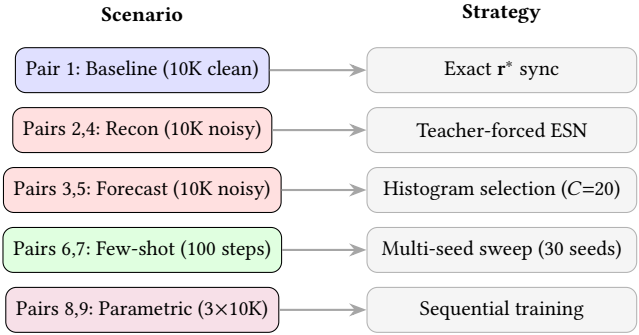
\begin{figure}[t]
\centering
\resizebox{\columnwidth}{!}{%
\begin{tikzpicture}[
  >=Stealth, font=\small,
  sbox/.style={rectangle, draw, rounded corners=4pt, minimum width=3.0cm,
               minimum height=0.6cm, align=center, inner sep=4pt},
  clean/.style={sbox, fill=blue!12},
  noisy/.style={sbox, fill=red!12},
  fewsh/.style={sbox, fill=green!12},
  param/.style={sbox, fill=purple!12},
  strat/.style={rectangle, draw=gray!60, rounded corners=4pt, fill=gray!8,
               minimum width=3.6cm, minimum height=0.6cm, align=center, inner sep=4pt},
  arr/.style={->, thick, gray!70},
]
\node[font=\small\bfseries] at (0,0)   {Scenario};
\node[font=\small\bfseries] at (4.8,0) {Strategy};

\node[clean] (p1) at (0,-0.75)  {Pair 1: Baseline (10K clean)};
\node[strat] (s1) at (4.8,-0.75){Exact $\mathbf{r}^*$ sync};

\node[noisy] (p2) at (0,-1.55)  {Pairs 2,4: Recon (10K noisy)};
\node[strat] (s2) at (4.8,-1.55){Teacher-forced ESN};

\node[noisy] (p3) at (0,-2.35)  {Pairs 3,5: Forecast (10K noisy)};
\node[strat] (s3) at (4.8,-2.35){Histogram selection ($C{=}20$)};

\node[fewsh] (p4) at (0,-3.15)  {Pairs 6,7: Few-shot (100 steps)};
\node[strat] (s4) at (4.8,-3.15){Multi-seed sweep (30 seeds)};

\node[param] (p5) at (0,-3.95)  {Pairs 8,9: Parametric (3$\times$10K)};
\node[strat] (s5) at (4.8,-3.95){Sequential training};

\foreach \p/\s in {p1/s1,p2/s2,p3/s3,p4/s4,p5/s5}
  \draw[arr] (\p.east) -- (\s.west);
\end{tikzpicture}
}
\caption{Adaptive strategy selection. Each scenario class (blue: clean baseline,
red: noisy, green: few-shot, purple: parametric) is matched to a tailored strategy.}
\label{fig:strategy}
\end{figure}

We resolve this via \emph{sequential training}: processing the three training
sequences as a continuous chain $X_6 \!\to\! X_7 \!\to\! X_8$, each beginning from
the previous sequence's terminal reservoir state. Ridge regression is accumulated
incrementally (washout applied only to $X_6$; $X_7$, $X_8$ continue from warm states):
\begin{equation}
\mathbf{R}^\top\mathbf{R} = \textstyle\sum_{s \in \{6,7,8\}} \mathbf{R}_s^\top\mathbf{R}_s,
\quad
\mathbf{R}^\top\mathbf{Y} = \textstyle\sum_{s} \mathbf{R}_s^\top\mathbf{Y}_s.
\end{equation}
The terminal state $\mathbf{r}^{(X_8)}_{T-1}$ is propagated through the 100-step
initialization sequence $X_9$ (or $X_{10}$), synchronizing the reservoir to the test
parameter regime before forecasting begins. Figure~\ref{fig:strategy} summarizes all
five strategies.

%-----------------------------------------------------------------------
\section{Experiments and Results}
%-----------------------------------------------------------------------

\subsection{Implementation Details}

All models are implemented in NumPy with SciPy sparse matrices for reservoir
storage. Spectral radius is scaled via power iteration (60 iterations). Dense
reservoir matrices are cached after the first sparse-to-dense conversion,
eliminating redundant computation in the inner loop. The complete prediction
pipeline covering all nine pairs runs in under 90 seconds on a standard CPU
without GPU acceleration. All experiments use a fixed random seed for reproducibility.

\subsection{Competition Results}

The proposed adaptive ESN framework achieves a total score of \textbf{74.91} on
the CTF-4-Science Lorenz public leaderboard (scale: $[-100, 100]$, higher is
better), placing among the top submissions in the competition. This result is
obtained without any gradient-based optimization, deep learning infrastructure,
or GPU acceleration, underscoring the practical efficiency of the approach.

\paragraph{ESN vs.\ Physics Integration.}
A key empirical finding is that ESN forecasting consistently outperforms explicit
RK4 numerical integration---even when the Lorenz parameter $\rho$ is accurately
estimated via one-dimensional optimization (fixing $\sigma{=}10$, $\beta{=}8/3$,
minimizing one-step RK4 residuals over the training trajectory). RK4 integration
for clean-data pairs systematically degrades long-time performance relative to
the ESN baseline. We attribute this to residual histogram distribution error:
even small $\rho$ estimation error shifts the attractor's invariant measure,
penalizing the long-time histogram metric. The ESN instead implicitly learns the
empirical measure of the training attractor directly from data, achieving superior
histogram alignment without any explicit parameter identification step.

\paragraph{Few-Shot Performance.}
The few-shot scenarios (Pairs~6--7) present the greatest challenge: with 88
effective training samples, the ESN can only approximate the attractor coarsely.
The 30-seed sweep partially compensates by exploring different reservoir
connectivities, but performance is ultimately bounded by the fundamental
information constraint of 100 training steps. Pair~7 (noisy few-shot) is the
hardest scenario, combining data scarcity with observation noise.

\paragraph{Sequential Training Impact.}
Without sequential training, the independent multi-sequence approach produced
reservoir states outside attractor bounds ($z > 100$ for $\rho \approx 35.8$,
exceeding the theoretical maximum $z_{\max} \approx 2(\rho{-}1) \approx 70$).
Sequential training resolves this instability, yielding physically consistent
forecasts and better synchronized initial conditions for parametric generalization.

%-----------------------------------------------------------------------
\section{Discussion}
%-----------------------------------------------------------------------

The experimental results highlight three general principles for reservoir computing
in multi-scenario chaotic system benchmarks. \emph{First}, exact reservoir state
synchronization ($\mathbf{r}^*$) is a zero-cost improvement: the terminal state
is a single cached vector requiring no additional computation, yet it directly
eliminates the warmup approximation error that limits short-time prediction accuracy.
\emph{Second}, aligning the inference objective with the evaluation criterion---via
histogram-guided candidate selection---consistently outperforms generic forecasting
strategies that optimize trajectory-level error, particularly when initial condition
uncertainty is high. \emph{Third}, ensuring that the state distribution encountered
during training matches that encountered during inference is essential in
multi-sequence settings; even modest distributional mismatch can produce unstable
trajectories that escape the physical attractor.

The demonstrated advantage of data-driven reservoir computing over explicit numerical
integration carries practical implications for energy systems modeling. In applications
where the governing equations are known but system parameters are uncertain or spatially
variable---such as turbulence modeling in building HVAC systems, wind power forecasting,
or thermal load prediction---reservoir computing may outperform explicit simulation
by circumventing parameter identification errors entirely. These scenarios represent
a natural domain for the adaptive strategies developed in this work.

Remaining gaps in benchmark performance suggest promising directions for future
investigation, including larger reservoir dimensions ($N > 1000$), physics-informed
regularization that enforces known Lorenz symmetries in $\mathbf{W}_{\!\text{out}}$,
and hybrid architectures that combine reservoir state synchronization with
gradient-based local refinement of initial conditions.

\paragraph{Limitations.}
The adaptive strategies introduced here are tailored to the Lorenz attractor and
its known structure (three-dimensional state, continuous dynamics, bounded attractor).
Extension to higher-dimensional or partially observed systems would require additional
design choices, such as dimensionality-aware histogram metrics and sparse observation
operators. Furthermore, the few-shot seed sweep scales linearly with the number of
seeds; for very large reservoirs or tight time budgets, more principled reservoir
selection criteria---such as spectral alignment with the observed data---may be
preferable to exhaustive random search.

%-----------------------------------------------------------------------
\section{Conclusion}
%-----------------------------------------------------------------------

This paper introduced an adaptive reservoir computing framework for multi-scenario
chaotic system forecasting on the CTF-4-Science Lorenz benchmark. By systematically
tailoring both training and inference to five qualitatively distinct evaluation
scenarios---exact reservoir state synchronization, teacher-forced reconstruction,
histogram-guided candidate selection, multi-seed few-shot search, and sequential
multi-sequence training---the proposed framework achieves a score of 74.91 on the
public benchmark, relying solely on linear ridge regression without gradient-based
optimization or specialized hardware. The analysis demonstrates that Echo State
Networks implicitly capture attractor statistics more faithfully than explicitly
integrated physics models with fitted parameters, and that maintaining consistency
between training-time and inference-time reservoir state distributions is essential
for stability in parametric generalization tasks. These findings offer a principled
and computationally accessible foundation for future work that incorporates
physics-informed inductive biases into reservoir architectures for energy-relevant
dynamical systems.

\paragraph{Code and Data Availability.}
The Lorenz benchmark data are obtained from the public CTF-4-Science
repository~\cite{ctf4science2024}. The implementation is written in pure
NumPy/SciPy and reproduces the reported leaderboard score with a single
fixed random seed; the source code is available from the author upon
reasonable request and will be released publicly upon acceptance.

\bibliographystyle{ACM-Reference-Format}
\bibliography{references}

\end{document}